\documentclass[11pt]{article}
\usepackage{eacl2017}
\usepackage[utf8]{inputenc}
\usepackage{times}
\usepackage{latexsym}
\usepackage{amsmath}
\usepackage{multirow}
\usepackage{url}
\usepackage{booktabs}
\usepackage{graphicx}
\usepackage{amssymb}
\usepackage{subfigure}
\usepackage{tikz}
\usepackage{pgfplots}
\usepackage{float}
\usepackage{tikz-dependency}
\usepackage{caption}

\eaclfinalcopy
\def\eaclpaperid{105}

\begin{document}
\sloppy
\title{Cross-Lingual Dependency Parsing with Late Decoding for Truly Low-Resource Languages}

\author{
Michael Sejr Schlichtkrull\\
  University of Amsterdam\thanks{\ Work done while at the University of Copenhagen.} \\
  {\tt m.s.schlichtkrull@uva.nl}  \And
  Anders Søgaard\\
   University of Copenhagen \\
  {\tt soegaard@di.ku.dk}}
\date{}
\maketitle

\begin{abstract}
In cross-lingual dependency annotation projection, information is often lost during transfer because of early decoding. We present an end-to-end graph-based neural network dependency parser that can be trained to reproduce matrices of edge scores, which can be directly projected across word alignments. We show that our approach to cross-lingual dependency parsing is not only simpler, but also achieves an absolute improvement of 2.25\% averaged across 10 languages compared to the previous state of the art.

\end{abstract}

\section{Introduction}\label{section:introduction}
Dependency parsing is an integral part of many natural language processing systems. However, most research into dependency parsing has focused on learning from treebanks, i.e. collections of manually annotated, well-formed syntactic trees. In this paper, we develop and evaluate a graph-based parser which does not require the training data to be well-formed trees. We show that such a parser has an important application in cross-lingual learning. 

Annotation projection is a method for developing parsers for low-resource languages, relying on aligned translations from resource-rich source languages into the target language, rather than linguistic resources such as treebanks or dictionaries. The Bible has been translated completely into 542 languages, and partially translated into a further 2344 languages. As such, the assumption that we have access to parallel Bible data, is much less constraining than the assumption of access to linguistic resources. Furthermore, for truly low-resource languages, relying upon the Bible scales better than relying on less biased data such as the EuroParl corpus.

In \newcite{agic2016parsing}, a projection scheme is proposed wherein labels are collected from many sources, projected into a target language, and then averaged. Crucially, the paper demonstrates how projecting and averaging edge scores from a graph-based parser \textit{before} decoding improves performance. Even so, decoding is still a requirement between projecting labels and retraining from the projected data, since their parser (TurboParser) requires well-formed input trees. This introduces a potential source of noise and loss of information that may be important for finding the best target sentence parse. 

Our approach circumvents the need for decoding prior to training, thereby surpassing a state-of-the-art dependency parser trained on decoded multi-source annotation projections as done by Agić et al. We first evaluate the model across several languages, demonstrating results comparable to the state of the art on the Universal Dependencies \cite{mcdonald2013universal} dataset. Then, we evaluate the same model by inducing labels from cross-lingual multi-source annotation projection, comparing the performance of a model with early decoding to a model with late decoding.

{\bf Contributions} We present a novel end-to-end neural graph-based dependency parser and apply it in a cross-lingual setting where the task is to induce models for truly low-resource languages, assuming only parallel Bible text. Our parser is more flexible than similar parsers, and accepts any weighted or non-weighted graph over a token sequence as input. In our setting, the input is a dense weighted graph, and we show that our parser is superior to previous best approaches to cross-lingual parsing. The code is made available on GitHub.\footnote{https://github.com/MichSchli/Tensor-LSTM}

\section{Model}\label{section:model}
The goal of this section is to construct a first-order graph-based dependency parser capable of learning \textit{directly} from potentially incomplete matrices of edge scores produced by another first-order graph-based parser. Our approach is to treat the encoding stage of the parser as a tensor transformation problem, wherein tensors of edge features are mapped to matrices of edge scores. This allows our model to approximate sets of scoring matrices generated by another parser directly through non-linear regression. The core component of the model is a layered sequence of recurrent neural network transformations applied to the axes of an input tensor.

\begin{figure}[h]
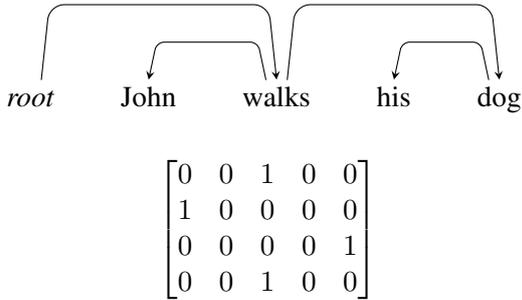

\centering     
\begin{dependency}[hide label]%[arc edge, arc angle=80] % Not supported by arXiv as they have not updated any packages since 2011.
\begin{deptext}[column sep=.7cm]
\textit{root} \& John \& walks \& his \& dog \\
\end{deptext}
\depedge{1}{3}{ }
\depedge{3}{2}{ }
\depedge{3}{5}{ }
\depedge{5}{4}{ }
\end{dependency}
\hfill
\begin{align*}
\begin{bmatrix}
0 & 0 & 1 & 0 & 0\\
1 & 0 & 0 & 0 & 0 \\
0 & 0 & 0 & 0 & 1 \\
0 & 0 & 1 & 0 & 0 \\
\end{bmatrix}
\end{align*}
\caption{An example dependency tree and the corresponding parse matrix.}
\label{figure:example-parse}
\end{figure}

More formally, any digraph $G = (V,E)$ can be expressed as a binary $|V| \times |V|$-matrix $M$, where $M_{ij} = 1$ if and only if $(j,i) \in E$ -- that is, if $i$ has an ingoing edge from $j$. If $G$ is a tree rooted at $v_0$, $v_0$ has no ingoing edges. Hence, it suffices to use a $(|V|-1) \times |V|$-matrix. In dependency parsing, every sentence is expressed as a matrix $S \in R^ {w \times f}$, where $w$ is the number of words in the sentence and $f$ is the width of a feature vector corresponding to each word. The goal is to learn a function $P: \mathbb{R}^{w \times f} \to \mathbb{Z}_2^{w \times (w +1)}$, such that $P(S)$ corresponds to the matrix representation of the correct parse tree for that sentence -- see Figure \ref{figure:example-parse} for an example.

In the arc-factored (first-order), graph-based model, $P$ is a composite function $P = D \circ E$ where the encoder $E : \mathbb{R}^{w \times f} \to \mathbb{R}^{w \times (w +1)}$ is a real-valued scoring function and the decoder $D: \mathbb{R}^{w \times (w +1)} \to \mathbb{Z}_2^{w \times (w +1)}$ is a minimum spanning tree algorithm \cite{mcdonald2005non}. Commonly, the encoder includes only \textit{local} information -- that is, $E_{ij}$ is only dependent on $S_i$ and $S_j$, where $S_i$ and $S_j$ are feature vectors corresponding to dependent and head.  Our contribution is the introduction of an LSTM-based \textit{global} encoder where the entirety of $S$ is represented in the calculation of $E_{ij}$.

We begin by extending $S$ to a $(w+1) \times (f+1)$-matrix $S^*$ with an additional row corresponding to the root node and a single binary feature denoting whether a node is the root.We now compute a 3-tensor $F = S \boxplus S^*$ of dimension $w \times (w+1) \times (2f+1)$ consisting of concatenations of all combinations of rows in $S$ and $S^*$. This tensor effectively contains a featurization of every edge $(u,v)$ in the complete digraph over the sentence, consisting of the features of the parent word $u$ and child word $v$. These edge-wise feature vectors are organized in the tensor exactly as the dependency arcs in a parse matrix such as the one shown in the example in Figure \ref{figure:example-parse}.

The edges represented by elements $F_{ij}$ can as such easily be interpreted in the context of related edges represented by the row $i$ and the column $j$ in which that edge occurs. The classical arc-factored parsing algorithm of \newcite{mcdonald2005non} corresponds to applying a function $O: \mathbb{R}^{2f+1} \to \mathbb{R}$ pointwise to $S \boxplus S^*$, then decoding the resulting $w \times (w+1)$-matrix. Our model diverges by applying an LSTM-based transformation $Q: \mathbb{R}^{w \times (w+1) \times (2f+1)} \to \mathbb{R}^{w \times (w+1) \times d}$ to $S \boxplus S^*$ before applying an analogous transformation $O: \mathbb{R}_{d} \to \mathbb{R}$.

The Long Short-Term Memory (LSTM) unit is a function $LSTM(x, h_{t-1}, c_{t-1}) = (h_t,c_t)$ defined through the use of several intermediary steps, following \newcite{hochreiter2001lstm}. A concatenated input vector $I = x \oplus h_{prev}$ is constructed, where $\oplus$ represents vector concatenation. Then, functions corresponding to input, forget, and output gates are defined following the form $g_{input} = \sigma (W_{input} I + b_{input})$. Finally, the internal cell state $c_t$ and the output vector $h_t$ at time $t$ are defined using the Hadamard (pointwise) product $\bullet$:
\begin{align*}
c_t &= g_{forget} \bullet c_{prev} + g_{input} \bullet tanh (W_{cell} I + b_{cell})\\
h_t &= g_{output} \bullet tanh(c_t)
\end{align*}

%Applying the commonly used Sequence-LSTM to the rows of a matrix $X$, 
We define a function Matrix-LSTM inductively, that applies an LSTM to the rows of a matrix $X$. Formally, Matrix-LSTM is a function $\mathcal{M}: \mathbb{R}^{a \times b}\to \mathbb{R}^{a \times c}$ such that $(h_1, c_1) = LSTM(X_1, 0, 0)$, $\forall 1 < i \leq n\ (h_i, c_i) = LSTM(X_i, h_{i-1}, c_{i-1})$, and $\mathcal{M}(X)_i = h_i$. 

\begin{figure*}[ht!]
\centering     
\includegraphics[scale=0.55]{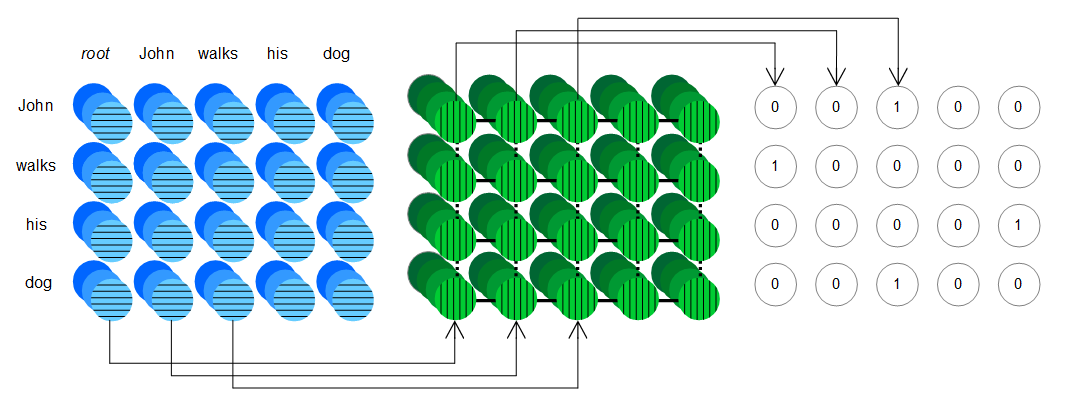}
\caption{Four-directional Tensor-LSTM applied to the example sentence seen in Figure \ref{figure:example-parse}. The word-pair tensor $S \boxplus S^*$ is represented with blue units (horizontal lines), a hidden Tensor-LSTM layer $H$ with green units (vertical lines), and the output layer with white units. The recurrent connections in the hidden layer along $H$ and $H^{T(2,1,3)}$ are illustrated respectively with dotted and fully drawn lines.}
\label{figure:tensor-lstm}
\end{figure*}

An effective extension is the \textit{bidirectional} LSTM, wherein the LSTM-function is applied to the sequence both in the forward and in the backward direction, and the results are concatenated. In the matrix formulation, reversing a sequence corresponds to inverting the order of the rows. This is most naturally accomplished through left-multiplication with an exchange matrix $J_m \in \mathbb{R}^{m \times m}$ such that:
\begin{align*}
J_m &=
\begin{bmatrix}
0 & \cdots & 1\\
\vdots & \reflectbox{$\ddots$} & \vdots \\
1 & \cdots &  0 \\
\end{bmatrix}
\end{align*}
Bidirectional Matrix-LSTM is therefore defined as a function $\mathcal{M}_{2d}: \mathbb{R}^{a \times b}\to \mathbb{R}^{a \times 2c}$ such that:
\begin{align*}
\mathcal{M}_{2d}(S) &= \mathcal{M}(S) \oplus_2 J_{a} \mathcal{M}(J_{a}S)
\end{align*}
Where $\oplus_2$ refers to concatenation along the second axis of the matrix.

Keeping in mind the goal of constructing a tensor transformation $Q$ capable of propagating information in an LSTM-like manner between any two elements of the input tensor, we are interested in constructing an equivalent of the Matrix-LSTM-model operating on 3-tensors rather than matrices. This construct, when applied to the edge tensor $F=S \boxplus S^*$, can then provide a means of interpreting edges in the context of related edges.

A very simple variant of such an LSTM-function operating on 3-tensors can be constructed by applying a bidirectional Matrix-LSTM to every matrix along the first axis of the tensor. This forms the center of our approach. Formally, bidirectional Tensor-LSTM is a function $\mathcal{T}_{2d}: \mathbb{R}^{a\times b\times c}\to\mathbb{R}^{a\times b\times 2h}$ such that:
\begin{align*}
\mathcal{T}_{2d}(T)_i &= \mathcal{M}_{2d}(T_i)
\end{align*}

This definition allows information to flow \textit{within} the matrices of the first axis of the tensor, but not \textit{between} them -- corresponding in Figure \ref{figure:tensor-lstm} to horizontal connection along the rows, but no vertical connections along the columns. To fully cover the tensor structure, we must extend this model to include connections along columns. 

This is accomplished through tensor transposition. Formally, tensor transposition is an operator $T^{T\sigma}$ where $\sigma$ is a permutation on the set $\{1,...,rank(T)\}$. The last axis of the tensor contains the feature representations, which we are not interested in scrambling. For the Matrix-LSTM, this leaves only one option -- $M^{T(1,2)}$. When the LSTM is operating on a 3-tensor, we have two options -- $T^{T(2,1,3)}$ and $T^{T(1,2,3)}$.  This leads to the following definition of four-directional Tensor-LSTM as a function $\mathcal{T}_{4d}: \mathbb{R}^{a\times b\times c}\to\mathbb{R}^{a\times b\times 4h}$ analogous to bi-directional Sequence-LSTMs:
\begin{align*}
\mathcal{T}_{4d}(T) &= \mathcal{T}_{2d}(T) \oplus_3\mathcal{T}_{2d}(T^{T(2,1,3)})^{T(2,1,3)}
\end{align*}

Calculating the LSTM-function on $T^{T(1,2,3)}$ and $T^{T(2,1,3)}$ can be thought of as constructing the recurrent links either "side-wards" or "downwards" in the tensor -- or, equivalently, constructing recurrent links either between the outgoing or between the in-going edges of every vertex in the dependency graph. In Figure \ref{figure:tensor-lstm}, we illustrate the two directions respectively with full or dotted edges in the hidden layer. 

The output of Tensor-LSTM is itself a tensor. In our experiments, we use a multi-layered variation implemented by stacking layers of models: $\mathcal{T}_{4d, stack}(T) = \mathcal{T}_{4d}(\mathcal{T}_{4d}(...\mathcal{T}_{4d}(T)...))$. We do not share parameters between stacked layers. Training the model is done by minimizing the value $\mathcal{E}(G, O(Q(S \boxplus S^*)))$ of some loss function $\mathcal{E}$ for each sentence $S$ with gold tensor $G$. We experiment with two loss functions. 

In our monolingual set-up, we exploit the  fact that parse matrices by virtue of depicting trees are right stochastic matrices. Following this observation, we constrain each row of $O(Q(S \boxplus S^*))$ under a softmax-function and use as loss the row-wise cross entropy. In our cross-lingual set-up, we use mean squared error. In both cases, prediction-time decoding is done with Chu-Liu-Edmonds algorithm \cite{edmonds1968optimum} following \newcite{mcdonald2005non}.

\section{Cross-lingual parsing}\label{section:multilingual}
\newcite{hwa2005bootstrapping} is a seminal paper for cross-lingual dependency parsing, but they use very detailed heuristics to ensure that the projected syntactic structures are well-formed.
\newcite{agic2016parsing} is the latest continuation of their work, presenting a new approach to cross-lingual projection, projecting edge scores rather than subtrees. \newcite{agic2016parsing} construct target-language treebanks by aggregating scores from multiple source languages, before decoding. Averaging before decoding is especially beneficial when the parallel data is of low quality, as the decoder introduces errors, when edge scores are missing. Despite averaging, there will still be scores missing from the input weight matrices, especially when the source and target languages are very distant. Below we show that we can circumvent error-inducing early decoding by training directly on the projected edge scores.

We assume source language datasets $\mathcal{L}_1, ..., \mathcal{L}_n$, parsed by monolingual arc-factored parsers, In our case, this data comes from the Bible. We assume access to a set of sentence alignment functions $A_s : \mathcal{L}_s \times \mathcal{L}_{t} \to \mathbb{R}_{0,1}$ where $A_s(S_{s},S_{t})$ is the confidence that $S_{t}$ is the translation of $S_{s}$. Similarly, we have access to a set of word alignment functions $W_{\mathcal{L}_{s}, S_s, S_t} : S_s \times S_t \to \mathbb{R}_{0,1}$ such that $S_s \in \mathcal{L}_s$, $S_t \in \mathcal{L}_{t}$, and $W(w_{s},w_{t})$ represents the confidence that $w_{s}$ aligns to $w_{t}$ given that $S_t$ is the translation of $S_s$

For each source language $\mathcal{L}_s$ with a scoring function $score_{\mathcal{L}_s}$, we define a local edge-wise voting function $vote_{S_{s}}((u_{s}, v_{s}), (u_{t}, v_{t}))$ operating on a source language edge $(u_s, v_s) \in S_s$ and a target language edge $(u_t, u_t) \in S_t$.  Intuitively, every source language edge votes for every target language edge with a score proportional to the confidence of the edges aligning and the score given in the source language. For every target language edge $(u_t, v_t) \in S_t$:
\begin{align*}
vote_{S_{s}}((u_{s}, v_{s}), (u_{t}, v_{t})) &= W_{\mathcal{L}_{s}, S_s, S_t}(u_{s}, u_{t})\\ &\ \ \ \cdot W_{\mathcal{L}_{s}, S_s, S_t}(v_{s}, v_{t})\\ &\ \ \ \cdot score_{\mathcal{L}_s}(u_s, v_s)
\end{align*}
Following \newcite{agic2016parsing}, a sentence-wise voting function is then constructed as the highest contribution from a source-language edge:
\begin{align*}
vote_{S_{s}}(u_{t}, v_{t}) &= \max\limits_{u_{s},v_{s} \in S_{s}} vote_{S_{s}}((u_{s}, v_{s}), (u_{t}, v_{t}))
\end{align*}
The final contribution of each source language dataset $\mathcal{L}_s$ to a target language edge $(u_t, v_t)$ is then calculated as the sum for all sentences $S_s \in \mathcal{L}_s$ over $vote_{S_s}(u_t, v_t)$ multiplied by the confidence that the source language sentence aligns with the target language sentence. For an edge $(u_t, v_t)$ in a target language sentence $S_{t} \in \mathcal{L}_{t}$:
\begin{align*}
vote_{\mathcal{L}_s}(u_t, v_t) &= \sum\limits_{S_s \in \mathcal{L}_s} A_s(S_{s}, S_{t})\ vote_{S_s}(u_t, v_t)
\end{align*}
Finally, we can compute a target language scoring function by summing over the votes for every source language:
\begin{align*}
score(u_t, v_t) &= \frac{\sum\limits_{i = 1}^n vote_{\mathcal{L}_i}(u_t, v_t)}{Z_{S_{t}}}
\end{align*}
Here, $Z_{S_{t}}$ is a normalization constant ensuring that the target-language scores are proportional to those created by the source-language scoring functions. As such, $Z_{S_{t}}$ should consist of the sum over the weights for each sentence contributing to the scoring function. We can compute this as:
\begin{align*}
Z_{S_{t}} &= \sum\limits_{i = 1}^n \sum\limits_{S_s \in L_i} A_s(S_s, S_{t})
\end{align*}
The sentence alignment function is not a probability distribution; it may be the case that no source-language sentences contribute to a target language sentence, causing the sum of the weights \textit{and} the sum of the votes to approach zero. In this case, we define $score(u_t, v_t) = 0$. Before projection, the source language scores are all standardized to have $0$ as the mean and $1$ as the standard deviation. Hence, this corresponds to assuming neither positive nor negative evidence concerning the edge.

We experiment with two methods of learning from the projected data -- decoding with Chu-Liu-Edmonds algorithm and then training as proposed in \newcite{agic2016parsing}, or directly learning to reproduce the matrices of edge scores. For alignment, we use the sentence-level \textit{hunalign} algorithm introduced in \newcite{vargaparallel} and the token-level model presented in \newcite{ostling2015bayesian}.

\section{Experiments}\label{section:experiments}
We conduct two sets of experiments. First, we evaluate the Tensor-LSTM-parser in the monolingual setting. We compare Tensor-LSTM to the TurboParser \cite{martins2010turbo} on several languages from the Universal Dependencies dataset. In the second experiment, we evaluate Tensor-LSTM in the cross-lingual setting. We include as baselines the delexicalized parser of \newcite{mcdonald2011multi}, and the approach of \newcite{agic2016parsing} using TurboParser. To demonstrate the effectiveness of circumventing the decoding step, we conduct the cross-lingual evaluation of Tensor-LSTM using cross entropy loss with \textit{early} decoding, and using mean squared loss with \textit{late} decoding.

\subsection{Model selection and training}
Our features consist of $500$-dimensional word embeddings trained on translations of the Bible. The word embeddings were trained using skipgram with negative sampling on a word-by-sentence PMI matrix induced from the Edinburgh Bible Corpus, following \cite{DBLP:journals/corr/LevySG16}. Our embeddings are not trainable, but fixed representations throughout the learning process. Unknown tokens were represented by zero-vectors. 

We combined the word embeddings with one-hot-encodings of POS-tags, projected across word alignments following the method of \newcite{agic2016parsing}. To verify the value of the POS-features, we conducted preliminary experiments on English development data. When including POS-tags, we found small, non-significant improvements for monolingual parsing, but significant improvements for cross-lingual parsing.

The weights were initialized using the normalized values suggested in \newcite{glorot2010understanding}. Following \newcite{jozefowicz2015empirical}, we add $1$ to the initial forget gate bias. We trained the network using RMSprop  \cite{tieleman2012lecture} with hyperparameters $\alpha = 0.1$ and $\gamma = 0.9$, using minibatches of 64 sentences. Following \newcite{neelakantan2015adding}, we added a noise factor $n \sim \mathcal{N}(0,\frac{1}{(1 + t)^{0.55}})$ to the gradient in each update. We applied dropouts after each LSTM-layer with a dropout probability $p=0.5$, and between the input layer and the first LSTM-layer with a dropout probability of $p=0.2$  \cite{bluche2015apply}. As proposed in \newcite{pascanu2012difficulty}, we employed a gradient clipping factor of $15$. In the monolingual setting, we used early stopping on the development set.

We experimented with 10, 50, 100, and 200 hidden units per layer, and with up to 6 layers. Using greedy search on monolingual parsing and evaluating on the English development data, we determined the optimal network shape to contain $100$ units per direction per hidden layer, and a total of $4$ layers.

\begin{figure}
\centering
\begin{tikzpicture}[scale=0.85]
\begin{axis}[
	xlabel=Epochs,
	ylabel=UAS,    
    legend style={at={(0.29,0.05)},anchor=south west}
    ]

\addplot[color=blue,mark=square] table[x=Iteration,y=Xent-5] {Data/5v10.dat};	
\addplot[color=red,mark=triangle] table[x=Iteration,y=L2-5] {Data/5v10.dat};
\addplot[color=green,mark=x] table[x=Iteration,y=Xent-10] {Data/5v10.dat};	
\addplot[color=cyan,mark=*] table[x=Iteration,y=L2-10] {Data/5v10.dat};

\legend{Cross entropy at 5000, Mean squared at 5000, Cross entropy at 10000, Mean squared at 10000}
\end{axis}
\end{tikzpicture}
\caption{UAS per epoch on German development data training from 5000 or 10000 randomly sampled sentences with projected annotations.}
\label{figure:5000-vs-10000}
\end{figure}
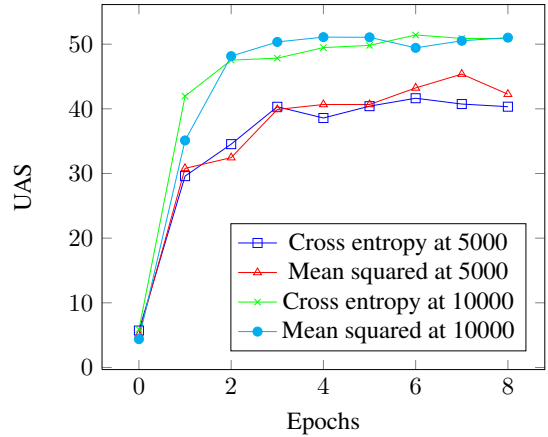

For the cross-lingual setting, we used two additional hyper-parameters. We used the development data from one of our target languages (German) to determine the optimal number of epochs before stopping. Furthermore, we trained only on a subset of the projected sentences, choosing the size of the subset using the development data. 

We experimented with either $5000$ or $10000$ randomly sampled sentences. There are two motivating factors behind this subsampling. First, while the Bible in general consists of about $30000$ sentences, for many low-resource languages we do not have access to annotation projections for the full Bible, because parts were never translated, and because of varying projection quality. Second, subsampling speeds up the training, which was necessary to make our experiments practical: At $10000$ sentences and on a single GPU, each epoch takes approximately $2.5$ hours. As such, training for a single language could be completed in less than a day. We plot the results in Figure \ref{figure:5000-vs-10000}. We see that the best performance is achieved at $10000$ sentences, and with respectively $6$ and $5$ epochs for cross entropy and mean squared loss.

\subsection{Results}\label{section:results}
In the monolingual setting, we compare our parser to TurboParser \cite{martins2010turbo} -- a fast, capable graph-based parser used as a component in many larger systems. TurboParser is also the system of choice for the cross-lingual pipeline of \newcite{agic2016parsing}. It is therefore interesting to make a direct comparison between the two. The results can be seen in Table \ref{table:results-monolingual}.

\begin{table}[htb]
\centering
\begin{tabular}{|l|r|r|r|}
\hline
Language        		&  TurboParser 	& Tensor-LSTM 	\\ \hline
English*				& 83.84		& \textbf{85.81}      	\\ \hline
German   		  	 	& 81.45		& \textbf{82.64}		    \\ \hline
Danish               	& 81.82		& \textbf{82.24}			\\ \hline
Finnish	        		& 77.74		& \textbf{78.83}			\\ \hline
Spanish         	 	& 83.19		& \textbf{86.69}			\\ \hline
French	        	 	& 81.17		& \textbf{84.63}			\\ \hline
Czech				  	& 81.32		& \textbf{85.04}			\\ \hline \midrule
Average 				& 81.50		& \textbf{83.70}				\\ \bottomrule
\end{tabular}
\caption{Unlabeled Attachment Score on the UD test data for TurboParser and Tensor-LSTM with cross entropy loss. English development data was used for model selection (marked *).}
\label{table:results-monolingual}
\end{table}

Note that in order for a parser to be directly applicable to the annotation projection setup explored in the secondary experiment, it must be a \textit{first-order graph-based} parser. In the monolingual setting, the best results reported so far ($84.74$, on average) for the above selection of treebanks were by the Parsito system \cite{straka2015baseline}, a transition-based parser using a dynamic oracle. 

For the cross-lingual annotation projection experiments, we use the delexicalized system suggested by \newcite{mcdonald2011multi} as a baseline. We also compare against the annotation projection scheme using TurboParser suggested in \newcite{agic2016parsing}, representing the previous state of the art for truly low-resource cross-lingual dependency parsing. Note that while our results for the TurboParser-based system use the same training data, test data, and model as in Agić et al., our results differ due to the use of the Bible corpus rather than a Watchtower publications corpus as parallel data. The authors made results available using the Edinburgh Bible Corpus for unlabeled data. The two tested conditions of Tensor-LSTM are the mean squared loss model \textit{without} intermediary decoding, and the cross entropy model \textit{with} intermediary decoding. The results of the cross-lingual experiment can be seen in Table \ref{table:results-crosslingual}.

\begin{table*}
\centering
\label{table:multilingual-results}
\begin{tabular}{|l|r|r|r|r|}
\hline
Language  		& Delexicalized & TurboParser & Tensor-LSTM & Tensor-LSTM	 \\ 
&			&  & (Decoding) & (No decoding)	 \\ \hline
Czech (cs)     	& 40.99         & {\bf 43.81}    & 42.58	& 41.54  \\ \hline
Danish (da)   	& 49.65         &  54.87    & \textbf{54.93} & 54.15  \\ \hline
English* (en)  	& 48.08         & 52.52          & \textbf{52.91}	& 52.90  \\ \hline
Finnish (fi)    & 41.18         & \textbf{46.08}          & 43.98	& 45.26  \\ \hline
French (fr)   	& 48.97         & 45.83          & \textbf{55.06}	& 53.83  \\ \hline
German* (de)    & 49.36         & 51.79          & \textbf{54.87}	& 53.85  \\ \hline
Spanish (es)  	& 47.60         & 58.90 		 & \textbf{59.60}	& 57.81  \\ \hline
Persian (fa)    & 28.93         & 14.88          & 46.47	& \textbf{48.60} \\ \hline
Hebrew (he)  	& 19.06         & {\bf 52.89} 	 & 26.17	& 31.41  \\ \hline
Hindi (hi)      & 21.03         & 43.31          & 43.21	& \textbf{46.09} \\ \hline
\midrule
Average&39.49&46.29&47.98&\textbf{48.54}\\
\bottomrule
\end{tabular}
\caption{Unlabeled attachment scores for the various systems. Tensor-LSTM is evaluated using cross entropy and mean squared loss. We include the results of two baselines -- the delexicalized system of \newcite{mcdonald2011multi} and the Turbo-based projection scheme of \newcite{agic2016parsing}. English and German development data was used for hyperparameter tuning (marked *).}
\label{table:results-crosslingual}
\end{table*}

\section{Discussion}\label{section:discussion}
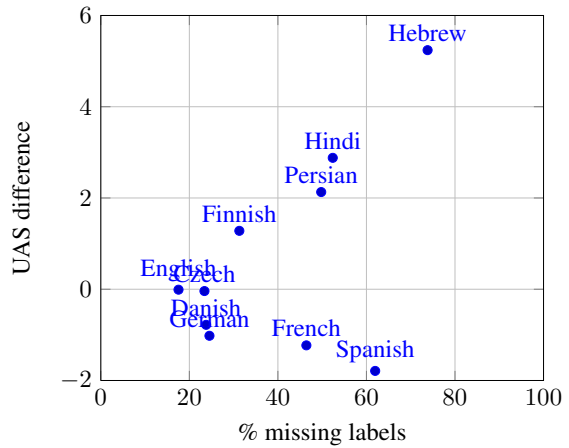
\begin{figure}[htb]
\centering
\begin{tikzpicture}[scale=0.85]
\begin{axis}[
  xmin=0, xmax=100, 
  ymin=-2, ymax=6,
  xlabel=\% missing labels,
  ylabel=UAS difference,
  grid=both,
  grid style={line width=.1pt, draw=gray!10},
  major grid style={line width=.2pt,draw=gray!50},
  ]
\addplot+[nodes near coords,
	only marks,
  	point meta=explicit symbolic,
    color=blue
    ]
  table[x=missing_percentage,y=diff,meta=language] {Data/missing_perf.dat};

\end{axis}
\end{tikzpicture}
\caption{Percentage of missing edge scores versus performance difference for Tensor-LSTM with mean squared loss and cross entropy loss.}
\label{figure:performance-difference}
\end{figure}

As is evident from Table \ref{table:results-crosslingual}, the variation in performance across different languages is large for all systems. This is to be expected, as the quality of the projected label sets vary widely due to linguistic differences. On average, Tensor-LSTM with mean squared loss outperforms all other systems. In Section \ref{section:introduction}, we hypothesized that incomplete projected scorings would have a larger impact upon systems reliant on an intermediary decoding step. To investigate this claim, we plot in Figure \ref{figure:performance-difference} the performance difference with mean squared loss and cross entropy loss for each language versus the percentage of missing edge scores.

For languages outside the Germanic and Latin families, our claim holds -- the performance of the cross entropy loss system decreases faster with the percentage of missing labels than the performance of the mean squared loss system. To an extent, this confirms our hypothesis, as we for the average language observe an improvement by circumventing the decoding step. French and Spanish, however, do not follow the same trend, with cross entropy loss outperforming mean squared loss despite the high number of missing labels.

In Table \ref{table:results-crosslingual}, performance on French and Spanish for both systems can be seen to be very high. It may be the case that indo-european target languages are not as affected by missing labels as most of the \textit{source} languages are themselves indo-european. Another explanation could be that some feature of the cross entropy loss function makes it especially well suited for Latin languages -- as seen in Table \ref{table:results-monolingual}, French and Spanish are also two of the languages for which Tensor-LSTM yields the highest performance improvement.

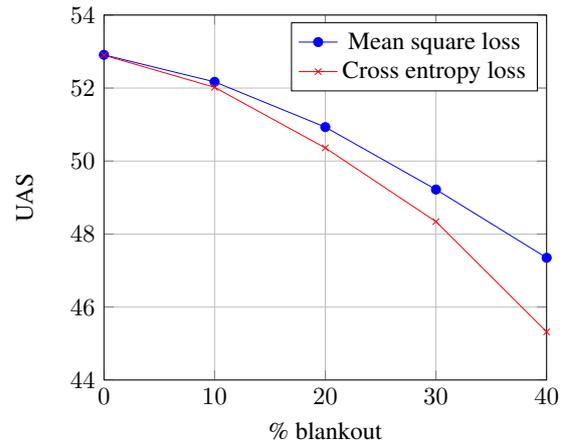
\begin{figure}[htb]
\centering
\begin{tikzpicture}[scale=0.85]
\begin{axis}[
  xmin=0, xmax=40, 
  ymin=44, ymax=54,
  xlabel=\% blankout,
  ylabel=UAS,
  grid=both,
  grid style={line width=.1pt, draw=gray!10},
  major grid style={line width=.2pt,draw=gray!50},
]
\addplot[
	color=blue, 
    mark=*
]
  table[x=blankout,y=uas_mse] {Data/blankout.dat};
  
\addplot[
	color=red, 
    mark=x
]
  table[x=blankout,y=uas_xent] {Data/blankout.dat};
  
 \legend{Mean square loss, Cross entropy loss}
\end{axis}
\end{tikzpicture}
\caption{Performance for Tensor-LSTM on English test data with $0$-$40 \%$ of the edge scores artificially maintained at $0$.}
\label{figure:blankout}
\end{figure}

To compare the effect of missing edge scores upon performance without influence from linguistic factors such as language similarity, we repeat the cross-lingual experiment on one language with respectively $10\%$, $20\%$, $30\%$, and $40\%$ of the projected and averaged edge scores artificially set to $0$, simulating missing data. We choose the English data for this experiment, as the English projected data has the lowest percentage of missing labels across any of the languages. In Figure \ref{figure:blankout}, we plot the performance for each of the two systems versus the percentage of deleted values.

As can be clearly seen, performance drops faster with the percentage of deleted labels for the cross entropy model. This confirms our intuition that the initially lower performance using mean squared loss compared to cross entropy loss is mitigated by a greater robustness towards missing labels, gained by circumventing the decoding step in the training process. In Table \ref{table:results-crosslingual}, this is reflected as dramatic performance increases using mean squared error for Finnish, Persian, Hindi, and Hebrew -- the four languages furthest removed from the predominantly indoeuropean source languages and therefore the four languages with the poorest projected label quality.

Several possible avenues for future work on this project are available. In this paper, we used an extremely simple feature function. More complex feature functions is one potential source of improvement. Another interesting direction for future work would be to include POS-tagging directly as a component of Tensor-LSTM prior to the construction of $S \boxplus S^*$ in a multi-task learning framework. Similarly, incorporating semantic tasks on top of dependency parsing could lead to interesting results. Finally, extensions of the Tensor-LSTM function to deeper models, wider models, or more connected models as seen in e.g. \newcite{kalchbrenner2015grid} may yield further performance gains.

\section{Related Work}\label{section:related}
Experiments with neural networks for dependency parsing have focused mostly on learning higher-order scoring functions and creating efficient feature representations, with the notable exception of \newcite{fonseca2015deep}. In their paper, a convolutional neural network is used to evaluate local edge scores based on global information. In \newcite{zhang2015high} and \newcite{pei2015effective}, neural networks are used to simultaneously evaluate first-order and higher-order scores for graph-based parsing, demonstrating good results. Bidirectional LSTM-models have been successfully applied to feature generation \cite{kiperwasser2016simple}. Such LSTM-based features could in future work be employed and trained in conjunction with Tensor-LSTM, incorporating global information both in parsing and in featurization.

An extension of LSTM to tensor-structured data has been explored in \newcite{graves2007mdlstm}, and further improved upon in \newcite{kalchbrenner2015grid} in the form of GridLSTM. Our approach is similar, but simpler and computationally more efficient as no within-layer connections between the first and the second axes of the tensor are required. 

Annotation projection for dependency parsing has been explored in a number of papers, starting with \newcite{hwa2005bootstrapping}. In \newcite{Tiedemann2014rediscovering} and \newcite{Tiedemann2015cross} the process in extended and evaluated across many languages. \newcite{li2014soft} follows the method of \newcite{hwa2005bootstrapping} and adds a probabilistic target-language classifier to determine and filter out high-uncertainty trees. In \newcite{ma2014unsupervised}, performance on projected data is used as an additional objective for unsupervised learning through a combined loss function.

A common thread in these papers is the use of high-quality parallel data such as the EuroParl corpus. For truly low-resource target languages, this setting is unrealistic as parallel resources may be restricted to biased data such as the Bible. In \newcite{agic2016parsing} this problem is addressed, and a parser is constructed which utilizes averaging over edge posteriors for many source languages to compensate for low-quality projected data. Our work builds upon their contribution by constructing a more flexible parser which can bypass a source of bias in their projected labels, and we therefore compared our results directly to theirs.

Annotation projection procedures for cross-lingual dependency parsing has been the focus of several other recent papers \cite{guo2015representations,Zhang2015hierarchical,Duong2015parameter,rasooli2015density}. In \newcite{guo2015representations}, distributed, language-independent feature representations are used to train shared parsers. \newcite{Zhang2015hierarchical} introduce a tensor-based feature representation capable of incorporating prior knowledge about feature interactions learned from source languages. In \newcite{Duong2015parameter}, a neural network parser is built wherein higher-level layers are shared between languages. 

Finally, \newcite{rasooli2015density} leverage dense information in high-quality sentence translations to improve performance. Their work can be seen as opposite to ours -- whereas Rasooli and Collins leverage high-quality translations to improve performance when such are available, we focus on improving performance in the \textit{absence} of high-quality translations.

\section{Conclusion}
We have introduced a novel algorithm for graph-based dependency parsing based on an extension of sequence-LSTM to the more general Tensor-LSTM. We have shown how the parser with a cross entropy loss function performs comparably to state of the art for monolingual parsing. Furthermore, we have demonstrated that the flexibility of our parser enables learning from non well-formed data and from the output of other parsers. Using this property, we have applied our parser to a cross-lingual annotation projection problem for truly low-resource languages, demonstrating an average target-language unlabeled attachment score of 48.54, which to the best of our knowledge are the best results yet for the task.

\section*{Acknowledgments}
The second author was supported by ERC Starting Grant
No. 313695.
\bibliographystyle{eacl2017}
\bibliography{refs}

\end{document}